\documentclass[journal]{IEEEtran}
\usepackage{amsmath}
\usepackage{algorithm}
\usepackage{algorithmic}
\usepackage{graphicx}
\usepackage{multirow}
\usepackage{tabularx}
\usepackage{xcolor}
\usepackage{array}
\usepackage{amssymb}
\usepackage{bbding}
\usepackage{subfigure}
\usepackage{diagbox}
\usepackage{rotating}
\usepackage{bm}
\usepackage{booktabs}
\usepackage{cite}
\usepackage{CJK}
\usepackage{overpic}
\usepackage[colorlinks,
            linkcolor=red,
            anchorcolor=blue,
            citecolor=green
            ]{hyperref}

\newcommand{\addFig}[1]{}
\newcommand{\addFigs}[1]{}

\usepackage{subfigure}
\newcommand{\etal}{\textit{et~al}.~}
\newcommand{\ie}{\textit{i}.\textit{e}.,~}
\newcommand{\eg}{\textit{e}.\textit{g}.,~}

\usepackage{balance}
\usepackage{cleveref}
\crefformat{section}{\S#2#1#3} 
\crefformat{subsection}{\S#2#1#3}
\crefformat{subsubsection}{\S#2#1#3}

\ifCLASSINFOpdf
\else
\fi

\begin{document}
%
\title{Multi-Content Complementation Network \\
for Salient Object Detection
\\in Optical Remote Sensing Images}
%
%
%

\author{Gongyang~Li,
	Zhi~Liu,~\IEEEmembership{Senior Member,~IEEE},
	Weisi~Lin,~\IEEEmembership{Fellow,~IEEE},
        and~Haibin~Ling
        
\thanks{Gongyang Li and Zhi Liu are with Shanghai Institute for Advanced Communication and Data Science, Shanghai University, Shanghai 200444, China, and School of Communication and Information Engineering, Shanghai University, Shanghai 200444, China.
Gongyang Li is also with the School of Computer Science and Engineering, Nanyang Technological University, Singapore 639798 (email: ligongyang@shu.edu.cn; liuzhisjtu@163.com).}
\thanks{Weisi Lin is with the School of Computer Science and Engineering, Nanyang Technological University, Singapore 639798 (e-mail: wslin@ntu.edu.sg).}
\thanks{Haibin Ling is with the Department of Computer Science, Stony Brook University, Stony Brook, NY 11794 USA (email: hling@cs.stonybrook.edu).}
\thanks{ \textit{Corresponding author: Zhi Liu.}}
}

\markboth{IEEE TRANSACTIONS ON GEOSCIENCE AND REMOTE SENSING}%
{Shell \MakeLowercase{\textit{et al.}}: Bare Demo of IEEEtran.cls for IEEE Journals}

\maketitle

\begin{abstract}
In the computer vision community, great progresses have been achieved in salient object detection from natural scene images (NSI-SOD); by contrast, salient object detection in optical remote sensing images (RSI-SOD) remains to be a challenging emerging topic.
The unique characteristics of optical RSIs, such as scales, illuminations and imaging orientations, bring significant differences between NSI-SOD and RSI-SOD.
In this paper, we propose a novel Multi-Content Complementation Network (MCCNet) to explore the complementarity of multiple content for RSI-SOD.
Specifically, MCCNet is based on the general encoder-decoder architecture, and contains a novel key component named Multi-Content Complementation Module (MCCM), which bridges the encoder and the decoder.
In MCCM, we consider multiple types of features that are critical to RSI-SOD, including foreground features, edge features, background features, and global image-level features, and exploit the content complementarity between them to highlight salient regions over various scales in RSI features through the attention mechanism.
Besides, we comprehensively introduce pixel-level, map-level and metric-aware losses in the training phase.
Extensive experiments on two popular datasets demonstrate that the proposed MCCNet outperforms 23 state-of-the-art methods, including both NSI-SOD and RSI-SOD methods.
The code and results of our method are available at https://github.com/MathLee/MCCNet.
\end{abstract}


\IEEEpeerreviewmaketitle

\section{Introduction}
\IEEEPARstart{V}{isual} attention mechanism aims to capture the most attractive regions in a scene, and plays an important role in the human visual system.
In computer vision, efforts have been devoted to model this mechanism and can be generally divided into two important topics: \emph{fixation prediction} and \emph{salient object detection}.
The former predicts visual saliency degree of regions, while the latter highlights salient object regions.
In this paper, we focus on salient object detection (SOD)~\cite{2015SODBenchmark,2019sodsurvey,WWG19Video,19CRMCO},
which has shown successful applications in various computer vision tasks, such as object segmentation~\cite{LGY2019,LGY2021PFOS}, image quality assessment~\cite{16SODIQA,19SGDNet}, image retargeting~\cite{2012Retargeting}, \textit{etc}.
And different from the classic SOD in natural scene images (NSI-SOD), we are dedicated to SOD in optical remote sensing images (RSI-SOD)~\cite{2019LVNet,2021DAFNet}.
Specifically, the optical RSIs refer to color images photographed by satellites and aerial sensors in the range of 400 to 760 nm~\cite{2019LVNet,2021RRNet}, and have only three optical bands (RGB), which are different from hyperspectral images that include more spectral bands information~\cite{R1-1}.
RSI-SOD aims at highlighting airplanes, islands, ships, buildings and rivers, which attract humans' attention, at the pixel level in the optical RSI.

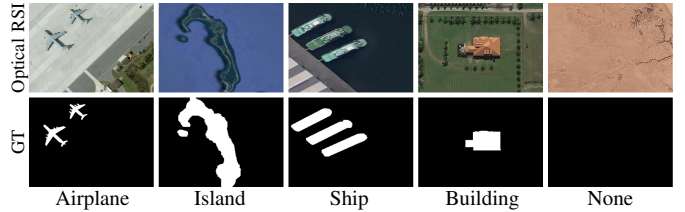
\begin{figure}[t!]
  \centering
  \footnotesize
  \begin{overpic}[width=0.99\columnwidth]{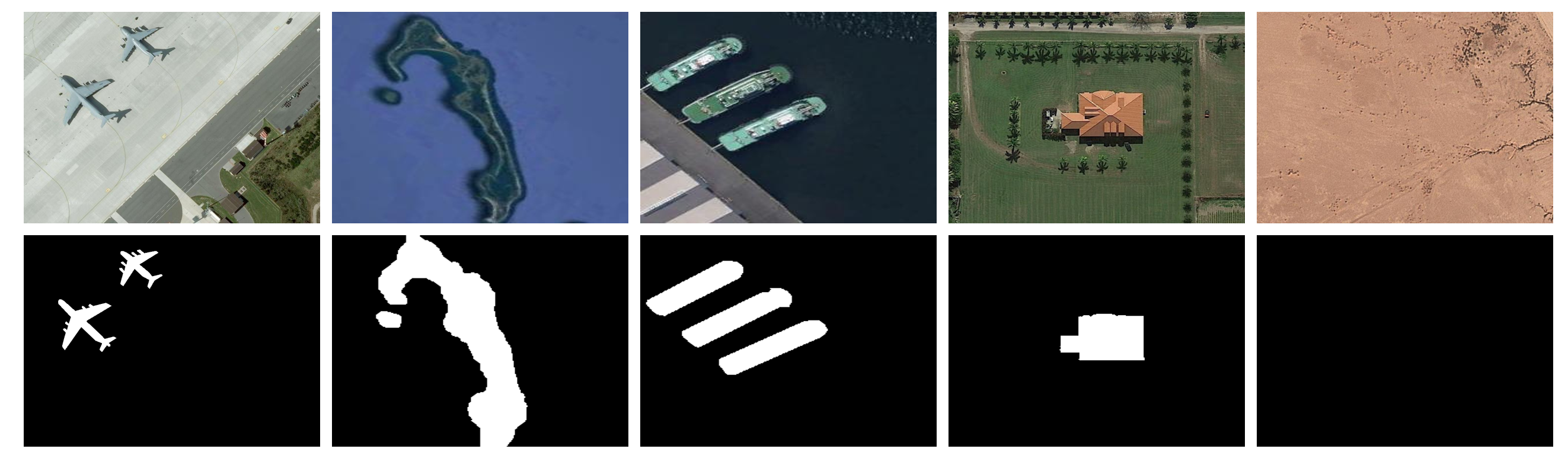}
    \put(4.4,-1.5){  {Airplane}  }
    \put(25.4,-1.5){ {Island} }
    \put(46.0,-1.5){ {Ship} }
    \put(63.5,-1.5){ {Building} }
    \put(85.05,-1.5){ {None} }
    \put(-2.3,15.55){ \begin{sideways}{\scriptsize{Optical RSI}}\end{sideways} }
    \put(-2.3,6.25){ \begin{sideways}{\scriptsize{GT}}\end{sideways} }
  \end{overpic}
  \caption{Representative example scenes in the RSI-SOD task. ``None'' means there is no salient object in this scene. GT is the ground truth.
    }\label{fig:example}
\end{figure}

%
Convolutional neural networks (CNNs)~\cite{1989CNN} significantly stimulate NSI-SOD~\cite{2019sodsurvey} and greatly improve the detection accuracy.
Recently, as many thought-provoking ideas and techniques, such as multi-level/scale fusion~\cite{2017Amulet}, edge guidance/preservation~\cite{2019EGNet,2018EdgeP}, attention~\cite{2020PiCANet,2020RANet}, complementary losses~\cite{2019BASNet,2019Floss}, \textit{etc}, are introduced into NSI-SOD, NSI-SOD has become more mature.
However, there are big differences between the acquisition of NSIs and optical RSIs.
Optical RSIs are photographed by satellite and aerial sensors, so the object types, scales, illuminations, imaging orientations and backgrounds of optical RSIs are fundamentally different from NSIs.
Some representative scenes in RSI-SOD task are shown in Fig.~\ref{fig:example}.
The last scene of Fig.~\ref{fig:example} is special, there is no salient object.
Thus, directly applying NSI-SOD methods to optical RSIs may be inappropriate.

However, as an emerging topic of saliency detection, RSI-SOD solutions are heavily inspired by NSI-SOD ones, especially the CNN-based ones.
Concretely, as a pioneer work in RSI-SOD, LVNet~\cite{2019LVNet} fuses multi-resolution inputs in a nested structure to perceive objects of different sizes.
PDFNet~\cite{2020PDFNet} integrates five-scale features from five branches for comprehensive detection.
DAFNet~\cite{2021DAFNet} not only employs the salient edge map as the additional supervision, but also performs attention in a dense fluid manner.
Similar to~\cite{2019LVNet}, EMFINet~\cite{2021EMFINet} adopts optical RSIs with three different resolutions as inputs,
but different from~\cite{2021DAFNet}, it employs edge supervision to generate features with edge-aware constraint
and introduces a hybrid loss to infer salient objects with shape boundaries.
These specialized CNN-based RSI-SOD methods are based on the characteristics of optical RSI to propose effective solutions and obtain promising performance.

Motivated by the above observations, we expand the advantages of NSI-SOD methods~\cite{2019EGNet,2018EdgeP,2020RANet} and propose a novel \emph{Multi-Content Complementation Module} (MCCM) to adapt to the characteristics of optical RSIs.
Specifically, we first integrate the foreground content into our MCCM.
Similar to~\cite{2019EGNet,2018EdgeP,2021DAFNet,2021EMFINet}, we introduce edge content, but the difference is that we employ edge supervision to produce an edge attention map for edge activation in features.
For RSI-SOD, we believe that in addition to foreground and edge, the background~\cite{2020RANet} is also important.
Here, we consider the complex background content of optical RSIs.
The above three kinds of content cover local information in detail.
Inspired by~\cite{DeeplabV3}, we incorporate global image-level content for comprehensive content complementation.
In this way, our MCCM captures both local and global content simultaneously, which is effective for accurately perceiving salient regions and distinguishing cluttered background regions.

Moreover, to improve the robustness of our MCCM, we implement MCCM at multiple feature scales.
We deploy MCCM in an encoder-decoder network which is a general backbone for NSI-SOD, and propose a simple yet effective \emph{Multi-Content Complementation Network} (MCCNet) for RSI-SOD. 
Benefiting from the progressive inference procedure in the backbone, our MCCNet can highlight salient regions with various scales and object types and flexibly adapt to the challenging scenes of optical RSIs.
In addition, following~\cite{2019BASNet,2019Floss}, we construct a comprehensive loss function to efficiently train our MCCNet.

Our main contributions are summarized as follows:
\begin{itemize}
\item We propose a \emph{Multi-Content Complementation Module} (MCCM) to explore the complementarity of multiple content in features of optical RSIs for salient regions perception.
In MCCM, the local content, \ie foreground, edge and background, and the global image-level content are simultaneously exploited.

\item We embed MCCM on multiple feature scales in an encoder-decoder network, and propose an effective and efficient \emph{Multi-Content Complementation Network} (MCCNet) for RSI-SOD, which runs at a fast inference speed of 95 \emph{fps} on a single GPU.
MCCNet perfectly combines the feature complementation ability of MCCM and the inference ability of the basic network.

\item We conduct comprehensive experiments on two benchmark RSI-SOD datasets.
The experimental results demonstrate that the proposed MCCNet is superior to 23 state-of-the-art methods under various evaluation metrics, and the effectiveness of the proposed MCCM is also verified.

\end{itemize}

The rest of this paper is organized as follows.
Sec.~\ref{sec:related} reviews the related work of NSI-SOD and RSI-SOD,
Sec.~\ref{sec:OurMethod} presents our MCCNet in details,
Sec.~\ref{sec:exp} elaborates experiments and ablation studies,
and Sec.~\ref{sec:con} draws the conclusion.

\section{Related Work}
\label{sec:related}
In this section, we first summarize the works of NSI-SOD, and then elaborate RSI-SOD methods.
For each topic, we introduce both traditional and CNN-based methods.

\subsection{Salient Object Detection in Natural Scene Images}
\label{sec:NSI_SOD}
As a pioneer in saliency detection, Itti \etal\cite{1998Itti} proposed the first computational visual attention model for NSIs, which is the cornerstone of other traditional work.
Liu~\etal\cite{Liu2012SOD} proposed an unsupervised method based on kernel density estimation.
In~\cite{Liu2014ST}, Liu~\etal proposed the saliency tree framework based on salient region merging and salient node selection.
The regularized random walks ranking was proposed in~\cite{2015RRWR}, and Yuan \etal\cite{2018RCRR} further combined it with reversion correction.
Meanwhile, Zou~\etal\cite{Liu2015SOD} jointly handled the SOD and object segmentation, effectively exploring the complementary cues of the two tasks.
In~\cite{2016HDCT}, Kim~\etal extended the high-dimensional color transform based SOD method with a local learning-based method.
Zhou~\etal\cite{2017DSG} integrated the diffusion results of foreground and background into the final saliency map.
In~\cite{2017SMD}, Peng~\etal applied the structured matrix decomposition to NSI-SOD.
Although traditional methods do not achieve impressive performance, they provide numerous valuable and thought-provoking solutions to NSI-SOD.

The CNN-based NSI-SOD methods~\cite{2019sodsurvey} break through the performance bottleneck of traditional methods~\cite{2015SODBenchmark} and promote NSI-SOD to a new era.
For instance, Hou~\etal\cite{2017DSS} implemented the deep supervision at multiple side-output layers for NSI-SOD.
Many subsequent methods~\cite{2020U2Net,2019EGNet,2020GCPA,2020ITSD,2020RANet,2021SUCA,2021PAKRN} have applied the deep supervision scheme to NSI-SOD.
Zhang~\etal\cite{2017Amulet} fused features over different scales to extract multi-scale information,
while Pang~\etal\cite{2020MINet} integrated features of three adjacent levels.
Zhao~\etal\cite{2020GateNet} proposed a gated dual branch control interference between different levels of features.
Edge/boundary cues were maturely used in NSI-SOD in various ways.
Wang~\etal\cite{2018EdgeP} directly extracted edge region from image, and sent it into the backbone network together with the image and superpixel region.
Differently, Wu~\etal\cite{2021DCN} used the Sobel operator to obtain the edge label as additional edge supervision.
Zhao~\etal\cite{2019EGNet} captured the salient edge from the ground truth and used it to force network learn edge features for one-to-one guidance.
Moreover, in~\cite{2020PiCANet}, Liu \etal learned pixel-wise local and global attention to facilitate detection.
Chen~\etal\cite{2020RANet} introduced the background information via the proposed reverse attention.
For supervision, in addition to the popular BCE loss, Ma~\etal\cite{2021PFS} introduced the IoU loss, Qin~\etal\cite{2019BASNet} further introduced the SSIM loss, and Xu~\etal\cite{2021PAKRN} introduced common losses of fixation prediction to NSI-SOD.
Zhao~\etal\cite{2019Floss} further proposed a metric-aware F-measure loss based on the popular evaluation metric F-measure~\cite{Fmeasure}.
Besides, the global context-aware aggregation~\cite{2019PoolNet,2020GCPA} and the recurrent mechanism~\cite{2018RADF,2018R3Net} have also been widely explored.

Though existing NSI-SOD methods cannot be directly applied to optical RSIs, they still provide important references for RSI-SOD.
Our method incorporates some advantages of these NSI-SOD methods, such as deep supervision, complementary losses, and edge information, to adapt to the particularity of optical RSIs.

\begin{figure*}
	\centering
	\begin{overpic}[width=1\textwidth]{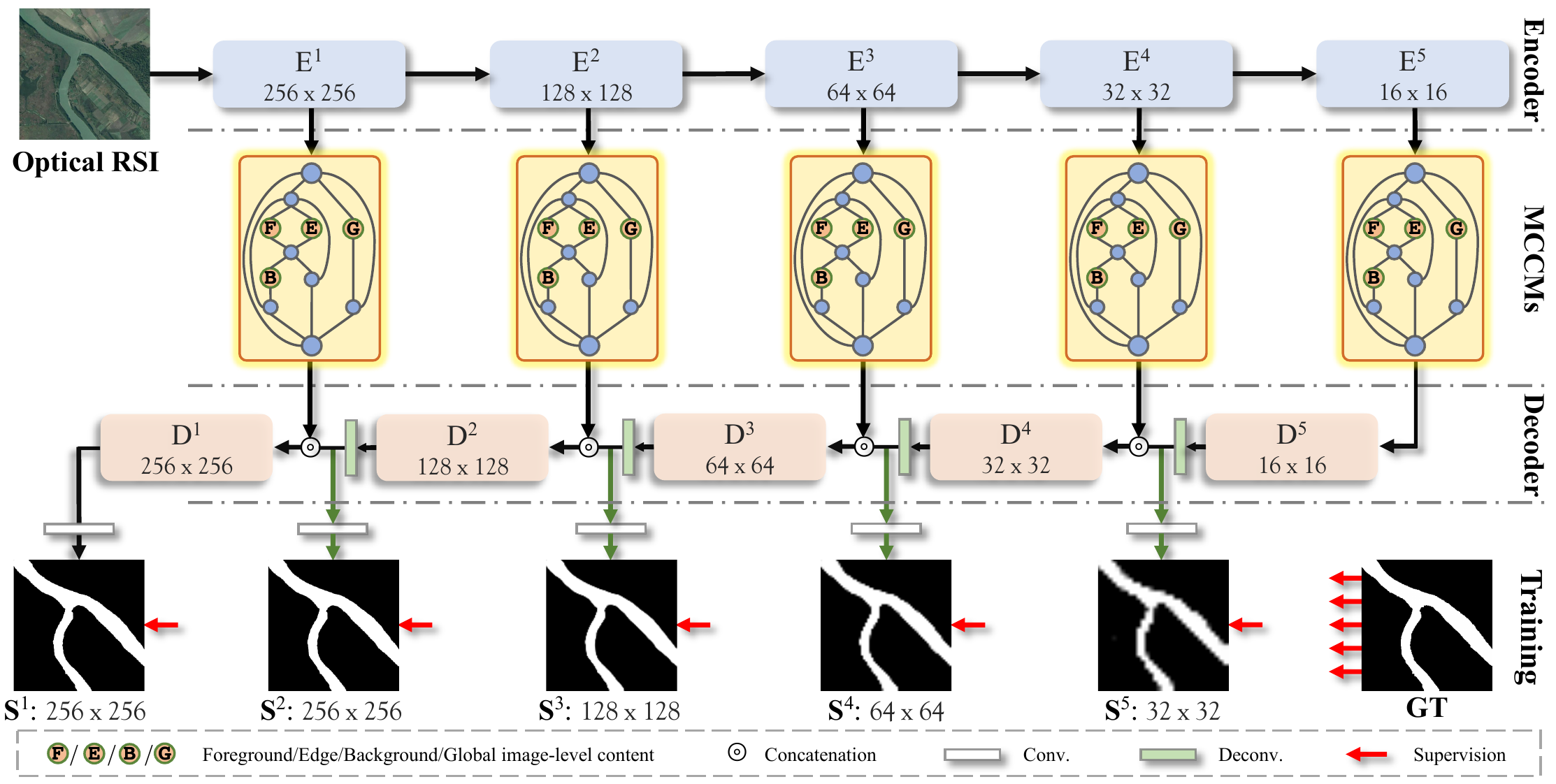}
    \end{overpic}
	\caption{The overall framework of the proposed MCCNet, which is based on the general encoder-decoder architecture.
	We first extract the basic features using the classic VGG-16~\cite{2014VGG16ICLR} from an optical RSI with a size of 256$\times$256$\times$3.
	Then, we model the complementary information between foreground features, edge features, background features and global image-level features in the pivotal Multi-Content Complementation Module (MCCM).
	Finally, we progressively infer salient objects using multi-scale features output from five MCCMs in the decoder.
	In the training phase, we employ the comprehensive supervision to each decoder block, including the pixel-level BCE loss, map-level IoU loss, and metric-aware F-m loss.
    }
    \label{fig:Framework}
\end{figure*}

\subsection{Salient Object Detection in Optical Remote Sensing Images}
\label{sec:ORSI_SOD}
Remote sensing image processing has been popular in the past decade.
Hong~\etal\cite{R1-2} proposed a general multimodal deep learning (MDL) framework, which consists of Ex-Net and Fu-Net, for pixel-level remote sensing image classification.
They introduced and developed five fusion architectures, including early fusion, middle fusion, late fusion, en-de fusion and cross fusion, in the MDL framework.
In~\cite{R1-3}, graph convolutional networks were introduced into hyperspectral image classification.
To address the shortage of identifying materials in cross-modality remote sensing data,
X-ModalNet\cite{R1-5}, a semi-supervised deep cross-modal framework, was proposed for classification in remote sensing data.
Moreover, Hong~\etal\cite{R1-4} proposed an augmented linear mixing model to address spectral variability for hyperspectral unmixing.
More in-depth analysis can be found in~\cite{R1-1}, which elaborates on the interpretable hyperspectral artificial intelligence.

In addition to the above popular tasks of remote sensing image processing,
there are some tasks similar to RSI-SOD,
such as airport detection~\cite{2018VOS},
ship detection~\cite{2018HSFNet,2019FHD,2019CR},
oil tank detection~\cite{2019CMC,2019SGSM},
building extraction~\cite{2017ISC},
residential areas extraction~\cite{2016GLSA,2021FDANet}, and
object detection from aerial images~\cite{ZhuWDBHL20arxiv}.
In fact, these object detection/extraction tasks mostly focus on specific scenes and objects, such as airport, ship, oil tank, building, and residential area.
By contrast, the RSI-SOD task involves all these scenarios, and is hence more general and challenging.

In particular, RSI-SOD task aims at extracting the most attractive objects in optical RSIs, and considers the subjective initiative of human more than the region-of-interest extraction task~\cite{2014FDA,2016SPS,2017SDBD,2019SD}.
Here, we first introduce some traditional RSI-SOD methods.
Faur~\etal\cite{2009RDM} regarded RSI-SOD as a data information compression task, and proposed a rate-distortion measure based method.
Based on the global and background information, Zhao~\etal\cite{2015SSD} proposed a sparse representation-based saliency computation method.
Zhang~\etal\cite{2015CIC,2018SPSS,2019SMFF} proposed a series of unsupervised methods:
in~\cite{2015CIC}, the saliency map was constructed based on color information content;
in~\cite{2018SPSS}, the statistical saliency feature map and the information saliency feature map were fused for final saliency map;
and in~\cite{2019SMFF}, a low-rank matrix recovery based self-adaptive multiple feature fusion method was proposed.
Compared with traditional RSI-SOD methods, CNN-based RSI-SOD methods provide more powerful solutions for complex optical RSIs.
In~\cite{2019LVNet} and \cite{2021DAFNet}, two challenging datasets of RSI-SOD were constructed.
And in~\cite{2019LVNet}, Li~\etal extracted multi-scale features directly from five different resolution optical RSIs in a two-stream pyramid module, and further perceived objects of different sizes in a V-shaped module with nested connections.
Following NSI-SOD methods like~\cite{2021DCN,2019EGNet}, Zhang~\etal\cite{2021DAFNet} constructed a multi-task architecture, which predicts saliency map and salient edge map simultaneously, to sharpen the object boundaries.
Furthermore, they proposed a cascaded pyramid attention module to solve the problem of object scale changes.
Following~\cite{2019LVNet}, Zhou~\etal\cite{2021EMFINet} extracted multi-scale features from three different resolution optical RSIs, and captured edge features via the edge supervision for edge preservation.
And they adopted the hybrid loss, including the pixel-level BCE loss, patch-level SSIM loss and map-level IoU loss~\cite{2019BASNet}, to facilitate the training.
Different from~\cite{2019LVNet,2021EMFINet}, Li~\etal\cite{2020PDFNet} only extracted features from a optical RSI, but efficiently captured multi-resolution information via integrating five different levels of features for inference.
Due to lack of optical RSIs data, Zhang~\etal\cite{2021PSL} introduced the weakly supervised learning into RSI-SOD.
They first generated pseudo labels with auxiliary images in a classification network, and then constructed a deep but lightweight feedback saliency analysis network to progressively refine saliency map.
We can find the impression of CNN-based NSI-SOD methods among the above specialized CNN-based RSI-SOD methods, but these specialized methods are uniquely constructed according to the characteristics of optical RSIs.

The above mentioned previous arts suggest that the foreground prior, edge clue and background cue play an important role in SOD.
However, they have been exploited independently in RSI-SOD.  
In our MCCM, we not only explore the complementation among three kinds of content (\ie foreground, edge and background), but also introduce the global (image-level) content, which is more comprehensive than~\cite{2021DAFNet} and~\cite{2021EMFINet}.
Moreover, we lay out MCCM with five feature scales in our MCCNet with one network input, which is more efficient than~\cite{2019LVNet} and~\cite{2021EMFINet}.

\section{Proposed Method}
\label{sec:OurMethod}
In this section, we detail the proposed Multi-Content Complementation Network.
In Sec.~\ref{sec:Overview}, we present the network overview of our MCCNet.
In Sec.~\ref{sec:MCCM}, we elaborate our Multi-Content Complementation Module (MCCM).
In Sec.~\ref{sec:Loss Function}, we clarify the comprehensive loss function.


\subsection{Network Overview}
\label{sec:Overview}
As depicted in Fig.~\ref{fig:Framework}, our MCCNet is built on the encoder-decoder architecture, which is friendly to pixel-level image segmentation~\cite{17SegNet,2015Unet} and various SOD tasks~\cite{2020GateNet,2019PoolNet,20ICNet,20CMWNet}, and is comprised of three key parts: the encoder network, five MCCM components and the decoder network.

For the encoder network, we adapt the popular VGG-16~\cite{2014VGG16ICLR} for basic feature extraction.
Different from the original VGG-16 structure for image classification task, we delete the last four layers, including one max-pooling layer and three fully connected layers, for our pixel-level RSI-SOD task, and denote the remaining five convolution blocks as E$^{t}$, where $\mathit{t}$ is the block index and belongs to $\{1, 2, 3, 4, 5\}$.
For $t=1,2$, E$^{t}$ contains two convolutional layers; for $t=3,4,5$, E$^{t}$ contains three convolutional layers.
For the input optical RSI $\boldsymbol{\mathrm{I}} \in \mathbb{R}^{256\!\times\!256\!\times\!3}$, the extracted features of each convolution block are denoted as $\boldsymbol{f}^{t}_\textrm{e} \in \mathbb{R}^{h_t\!\times\!w_t\!\times\!c_t}$, where $h_t $ is $\frac{256}{2^{t-1}}$, $w_t$ is $\frac{256}{2^{t-1}}$, and $c_t$ belongs to $\{64,128,256,512,512\}$.
Then, the basic features $\boldsymbol{f}^{t}_\textrm{e}$ of five levels will be fed to the corresponding MCCM to produce $\boldsymbol{f}^{t}_\textrm{mccm} \in \mathbb{R}^{h_t\!\times\!w_t\!\times\!c_t}$.
In MCCM, we generate the foreground, edge, background and global image-level features from the source features $\boldsymbol{f}^{t}_\textrm{e}$, and explore the complementarity between them.
Since MCCMs are deployed in five levels, they can capture multi-scale complementary information, which is beneficial to RSI-SOD.
Finally, our decoder network infers salient objects based on $\boldsymbol{f}^{t}_\textrm{mccm}$ in a progressive resolution restoration manner.
Our decoder network also contains five blocks, denoted by D$^{t}$, whose structure corresponds to that of E$^{t}$, \ie D$^{t}$ contains two convolutional layers for $t=1,2$, and D$^{t}$ contains three convolutional layers for $t=3,4,5$.
Between the two decoder blocks, we use a deconvolutional layer to restore the resolution.
In addition to the classic binary cross-entropy (BCE) loss, we introduce intersection-over-union (IoU) loss and F-measure (F-m) loss as auxiliary losses for each decoder block to comprehensively supervise the network training.
%

\begin{figure}
\centering
\footnotesize
  \begin{overpic}[width=1\columnwidth]{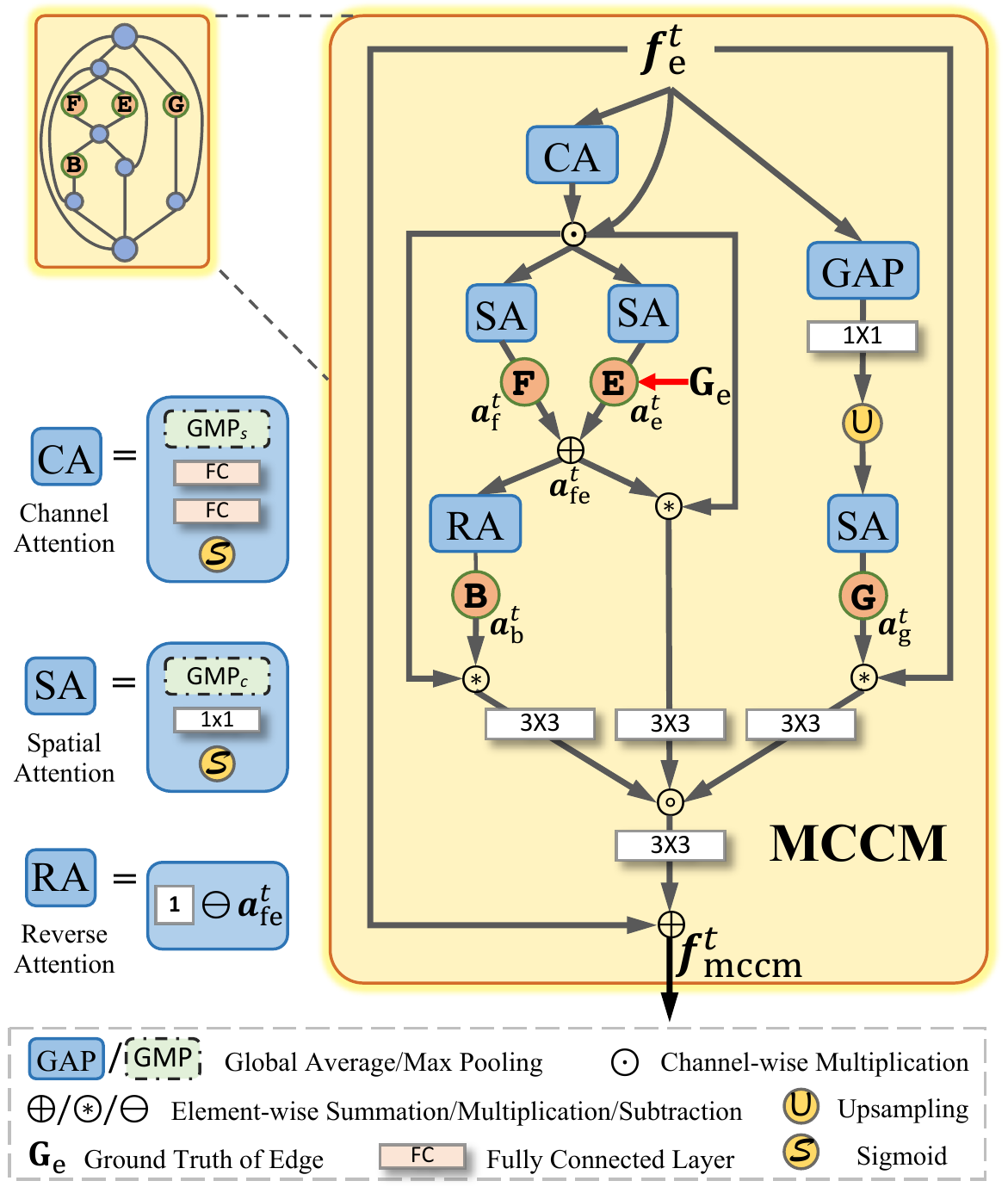}
  \end{overpic}
\caption{
Illustration of the Multi-Content Complementation Module (MCCM).
}
\label{MCCM_structure}
\end{figure}

\subsection{Multi-Content Complementation Module}
\label{sec:MCCM} 
In the studies of saliency detection, foreground assumption is often regarded as the prior in traditional NSI-SOD methods~\cite{2017DSG,2015RRWR}, and provides effective guidance to find obvious salient regions.
It can be explored in deep learning via the attention mechanism~\cite{SENet}.
The background information is another important cue in NSI-SOD to determine the non-salient regions, and is modeled by the reverse attention~\cite{2020RANet}.
The edge information is also widely used to complete the salient objects.
Since the scenes of RSI-SOD are typically more complicated than that of NSI-SOD, it is insufficient to consider the above three kinds of content in RSI-SOD independently.
According to the above motivation, we propose the \emph{Multi-Content Complementation Module} (MCCM), so as to address RSI-SOD based on the complementation of foreground, background and edge.
In addition, we integrate the global image-level content into MCCM, which is indispensable for RSI-SOD.
%
%

%
We illustrate the structure of MCCM in Fig.~\ref{MCCM_structure}.
In MCCM, the input features are $\boldsymbol{f}^{t}_\textrm{e}$ from E$^t$.
We regard $\boldsymbol{f}^{t}_\textrm{e}$ as source features, which generate all four kinds of content.
In the following, we elaborate MCCM based on these four kinds of content.

\textit{1) Foreground and Edge}.
As shown in Fig.~\ref{MCCM_structure}, foreground and edge features are extracted in parallel.
Considering that $\boldsymbol{f}^{t}_\textrm{e}$ of VGG-16 is relatively rough, we first perform the channel attention~\cite{SENet} on $\boldsymbol{f}^{t}_\textrm{e}$ to reduce redundant information and purify $\boldsymbol{f}^{t}_\textrm{e}$ as follows:
\begin{equation}
   \begin{aligned}
    \boldsymbol{f}^{t}_\textrm{ca} =   \mathrm{CA}(\boldsymbol{f}^{t}_\textrm{e}) \odot \boldsymbol{f}^{t}_\textrm{e},
    \label{eq:1}
    \end{aligned}
\end{equation}
where $\boldsymbol{f}^{t}_\textrm{ca} \in \mathbb{R}^{h_t\!\times\!w_t\!\times\!c_t}$ denotes the purified features,
$ \mathrm{CA}(\cdot)$ is the channel attention\footnote{Channel attention is implemented by a spatial global max pooling (GMP$_s$), two fully connected layers and a sigmoid activation function.}, and
$\odot$ is the channel-wise multiplication.

Then, 
we obtain the foreground map and edge map, denoted by
$\{\boldsymbol{a}^{t}_\textrm{f}, \boldsymbol{a}^{t}_\textrm{e}\}\in [0, 1]^{h_t\!\times\!w_t\!\times\!1}$, through the spatial attention~\cite{SENet} at the same time,
which can be computed as:
\begin{equation}
   \begin{aligned}
    \boldsymbol{a}^{t}_\textrm{f} =   \mathrm{SA}(\boldsymbol{f}^{t}_\textrm{ca}),
    \label{eq:2}
    \end{aligned}
\end{equation}
\begin{equation}
   \begin{aligned}
    \boldsymbol{a}^{t}_\textrm{e} =   \mathrm{SA}(\boldsymbol{f}^{t}_\textrm{ca}),
    \label{eq:3}
    \end{aligned}
\end{equation}
where $\mathrm{SA}(\cdot)$ is the spatial attention\footnote{Spatial attention is implemented by a channel global max pooling (GMP$_c$), a convolutional layer and a sigmoid activation function.}.
Notably, the foreground map is generated in an adaptive way, while the edge map is generated in a learning way, \ie under the supervision of the ground truth of edge in the training phase.

Since both foreground map and edge map are correlated to the salient regions and can complement each other, we aggregate them together via the 
element-wise summation, and obtain the foreground-edge map, denoted by 
$\boldsymbol{a}^{t}_\textrm{fe} \in [0, 2]^{h_t\!\times\!w_t\!\times\!1}$.
And we adopt the foreground-edge map to highlight the salient regions at feature level as follows:
\begin{equation}
   \begin{aligned}
       \boldsymbol{f}^{t}_\textrm{fe} =  \boldsymbol{a}^{t}_\textrm{fe} \circledast \boldsymbol{f}^{t}_\textrm{ca},
    \label{eq:4}
    \end{aligned}
\end{equation}
where $\boldsymbol{f}^{t}_\textrm{fe} \in \mathbb{R}^{h_t\!\times\!w_t\!\times\!c_t}$ is the foreground-edge features and $\circledast$ is the element-wise multiplication.
The way we merge the foreground map and edge map is different from those of using edge information in SOD~\cite{2019EGNet,2019BASNet,2019LVNet,2021EMFINet}.
In particular, our method explicitly explores the complementary information between these two maps, and is therefore more effective.

\textit{2) Background}.
The generation of background map is closely related to the foreground-edge map.
Following~\cite{2020RANet}, we obtain the background map, denoted by 
$\boldsymbol{a}^{t}_\textrm{b} \in [-1, 1]^{h_t\!\times\!w_t\!\times\!1}$, through the reverse attention.
In the background map $\boldsymbol{a}^{t}_\textrm{b}$, we redefine the concept of background and foreground, \ie the background is defined as 1, and the foreground is defined as -1.
And we also adopt the background map to highlight the non-salient regions at feature level.
The process is written as:
\begin{equation}
   \begin{aligned}
       \boldsymbol{a}^{t}_\textrm{b} =   \textbf{1} \ominus \boldsymbol{a}^{t}_\textrm{fe},
    \label{eq:5}
    \end{aligned}
\end{equation}
\begin{equation}
   \begin{aligned}
       \boldsymbol{f}^{t}_\textrm{b} =  \boldsymbol{a}^{t}_\textrm{b} \circledast \boldsymbol{f}^{t}_\textrm{ca},
    \label{eq:6}
    \end{aligned}
\end{equation}
where \textbf{1} is a matrix with size $h_t\!\times\!w_t\!\times\!1$, where all elements are 1, $\ominus$ is the element-wise subtraction, and $\boldsymbol{f}^{t}_\textrm{b} \in \mathbb{R}^{h_t\!\times\!w_t\!\times\!c_t}$ is the background features.
We can find that the background features are based on the foreground-edge features, but they are the opposite of the foreground-edge features.
In the subsequent processing of MCCM, we will merge them at channel level, and implicitly extract the complementary information between them.

\textit{3) Global Image-level Content}.
In fact, whether it is foreground-edge features or background features, they contain local information, which benefits to complete the detail and boundary of salient objects.
And inspired by~\cite{DeeplabV3}, we introduce the global image-level content to capture the overall tone of source features in our MCCM.

Concretely, following~\cite{DeeplabV3}, we apply spatial-wise global average pooling on $\boldsymbol{f}^{t}_\textrm{e}$ to extremely compress global distribution information into pixels and get the basic image-level features, and perform a 1$\times$1 convolutional layer for feature smoothing.
Then, we reconstruct the image-level content to the same size as the original $\boldsymbol{f}^{t}_\textrm{e}$ via the upsampling with bilinear interpolation.
Such a rough operation will lose a lot of detailed information, but the reconstructed features can reflect the overall tone of source features.
Different from~\cite{DeeplabV3} which directly integrates the image-level content at channel level via the concatenation, we compress the reconstructed image-level content into an elegant response map, namely the global image-level map
$\boldsymbol{a}^{t}_\textrm{g} \in [0, 1]^{h_t\!\times\!w_t\!\times\!1}$, via the spatial attention.
The entire process is formulated as follows:
\begin{equation}
   \begin{aligned}
       \boldsymbol{a}^{t}_\textrm{g} =  \mathrm{SA} \Big(\mathrm{up} \big( \mathrm{conv}_{1\times1} ( \mathrm{GAP}_{s} ( \boldsymbol{f}^{t}_\textrm{e} ) ) \big) \Big),
    \label{eq:7}
    \end{aligned}
\end{equation}
where $\mathrm{GAP}_{s}(\cdot)$ is the spatial global average pooling,
$\mathrm{conv}_{1\times1}(\cdot)$ is the 1$\times$1 convolutional layer, and
$\mathrm{up} (\cdot)$ is the upsampling operation.
We adopt $\boldsymbol{a}^{t}_\textrm{g}$ to reflect the overall tone at feature level as follows:
\begin{equation}
   \begin{aligned}
       \boldsymbol{f}^{t}_\textrm{g} =  \boldsymbol{a}^{t}_\textrm{g} \circledast \boldsymbol{f}^{t}_\textrm{e},
    \label{eq:8}
    \end{aligned}
\end{equation}
where $\boldsymbol{f}^{t}_\textrm{g} \in \mathbb{R}^{h_t\!\times\!w_t\!\times\!c_t}$ is the global image-level features.

\begin{figure}
\centering
\footnotesize
  \begin{overpic}[width=1\columnwidth]{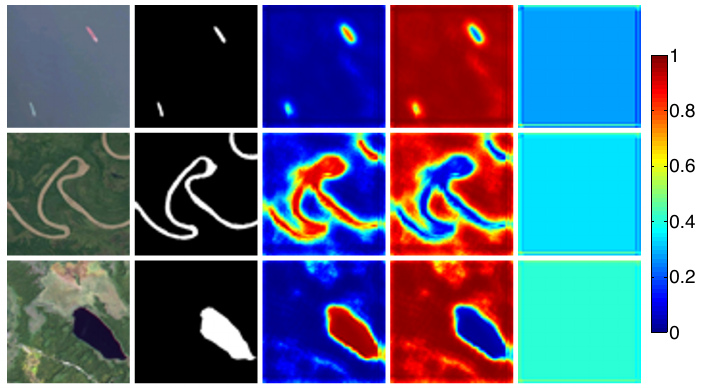}
    \put(0.97,-2.1){ {Optical RSI} }
    \put(24.45,-2.1){ {GT} }
    \put(41.9,-2.3){ { $\boldsymbol{a}^{\textrm{3}}_\textrm{fe}$  } }
    \put(59.9,-2.3){ { $\boldsymbol{a}^{\textrm{3}}_\textrm{b}$ }  }
    \put(78.05,-2.3){ { $\boldsymbol{a}^{\textrm{3}}_{\textrm{g}}$  }}
  \end{overpic}
\caption{
Feature visualization of components in MCCM attached to E$^\textrm{3}$.
Please zoom-in for viewing details.
}
\label{feature_visual}
\end{figure}
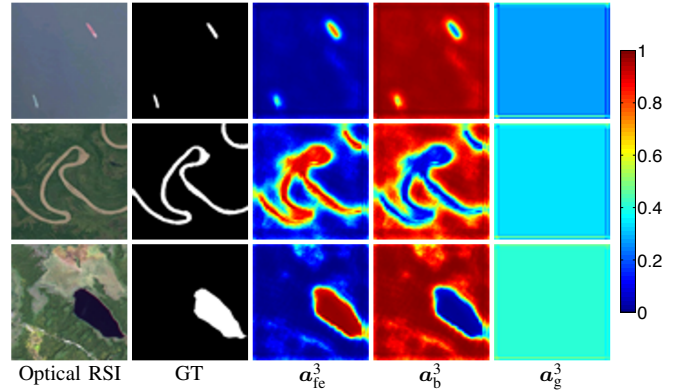

%
\textit{4) Multi-Content Aggregation.}
Through the above thorough operations, we obtain features of four kinds of content, \ie $\boldsymbol{f}^{t}_\textrm{fe}$, $\boldsymbol{f}^{t}_\textrm{b}$ and  $\boldsymbol{f}^{t}_\textrm{g}$, and further polish them via the 3$\times$3 convolutional layer, obtaining $\boldsymbol{\hat{f}}^{t}_\textrm{fe}$, $\boldsymbol{\hat{f}}^{t}_\textrm{b}$ and $\boldsymbol{\hat{f}}^{t}_\textrm{g}$.
Then, we aggregate them using the adaptive concatenation-convolution operation.
Besides, we adopt a short connection to retain the original content to generate the output features of MCCM $\boldsymbol{f}^{t}_\textrm{mccm} \in \mathbb{R}^{h_t\!\times\!w_t\!\times\!c_t}$.
The entire aggregation process is written as:
\begin{equation}
   \begin{aligned}
       \boldsymbol{f}^{t}_\textrm{mccm} =  \mathrm{conv}_{3\times3} ( \boldsymbol{\hat{f}}^{t}_\textrm{fe} \circledcirc \boldsymbol{\hat{f}}^{t}_\textrm{b}  \circledcirc  \boldsymbol{\hat{f}}^{t}_\textrm{g})  \oplus \boldsymbol{f}^{t}_\textrm{e},
    \label{eq:9}
    \end{aligned}
\end{equation}
where $\circledcirc$ is the cross-channel concatenation and $\oplus$ is the element-wise summation.
In summary, $\boldsymbol{f}^{t}_\textrm{mccm}$ is comprised of the essence of foreground, edge, background, global image-level content and original content, which makes MCCM an indispensable part of MCCNet.
The MCCMs equipped with five feature levels assist MCCNet to adapt to the complex scenes and changeable objects in optical RSIs.

In Fig.~\ref{feature_visual}, we visualize three maps of four kinds of content in MCCM attached to E$^\textrm{3}$, \ie $\boldsymbol{a}^{\textrm{3}}_\textrm{fe}$, $ \boldsymbol{a}^{\textrm{3}}_\textrm{b}$ and $\boldsymbol{a}^{\textrm{3}}_\textrm{g}$.
With the combination of $\boldsymbol{a}^{\textrm{3}}_\textrm{f}$ and $\boldsymbol{a}^{\textrm{3}}_\textrm{e}$, $\boldsymbol{a}^{\textrm{3}}_\textrm{fe}$ can clearly highlight the salient regions, regardless of whether they have a tiny size, a large size, multiple objects or a complex topology, and meanwhile $ \boldsymbol{a}^{\textrm{3}}_\textrm{b}$ makes the non-salient regions obvious.
And $\boldsymbol{a}^{\textrm{3}}_\textrm{g}$ provides the specific basic tone of source features of each optical RSI.

\subsection{Comprehensive Loss Function}
\label{sec:Loss Function}
For a successful CNN-based model, an effective architecture and several well-designed modules are necessary.
In addition, a good training strategy can improve model performance without extra model parameters.
As shown in Fig.~\ref{fig:Framework}, we adopt the widely used deep supervision~\cite{2015DeepSup,2017DSS} in the training phase to monitor the intermediate saliency maps of different sizes, which forces features to learn the characteristics of salient regions of different sizes.
And inspired by the successful usage of hybrid and complementary loss in SOD~\cite{2019BASNet,21HAINet,2021EMFINet}, we adopt the classic pixel-level BCE loss and map-level IoU loss in our loss function.
We also include the metric-aware F-m loss~\cite{2019Floss} in our loss function to further facilitate the network training.
Thus, we construct a comprehensive loss function $\mathbb{L}^{t}_{\mathrm{s}}$ to supervise the predicted saliency map $\mathbf{S}^{t}$ of D$^t$, which can be formulated as:
\begin{equation}
   \begin{aligned}
    \mathbb{L}^{t}_{\mathrm{s}}  = \ell_{bce} (\mathrm{up}(\mathbf{S}^{t}),\mathbf{G}) + \ell_{iou} (\mathrm{up}(\mathbf{S}^{t}),\mathbf{G}) + \ell_{fm} (\mathrm{up}(\mathbf{S}^{t}),\mathbf{G}),
    \label{eq:SalLoss}
    \end{aligned}
\end{equation}
where $\mathbf{G}$ is the ground truth, and $\ell_{bce} (\cdot)$, $\ell_{iou} (\cdot)$ and $\ell_{fm} (\cdot)$ are BCE loss, IoU loss and F-m loss, respectively.
They can be computed as follows:
\begin{equation}
   \begin{aligned}
    \ell_{bce}  = 
    -\sum\nolimits_{i=1}^{W \cdot H} [ \mathbf{G}(i) \mathrm{log}(\mathbf{S}(i)) 
    + (1-\mathbf{G}(i)) \mathrm{log}(1-\mathbf{S}(i)) ] ,
    \label{eq:BCELoss}
    \end{aligned}
\end{equation}
\begin{equation}
   \begin{aligned}
    \ell_{iou}  = 
    1 - \frac{  \sum\nolimits_{i=1}^{W \cdot H} \mathbf{S}(i) \cdot \mathbf{G}(i) }
    { \sum\nolimits_{i=1}^{W \cdot H} [\mathbf{S}(i) + \mathbf{G}(i) - \mathbf{S}(i) \cdot \mathbf{G}(i) ]},
    \label{eq:IoULoss}
    \end{aligned}
\end{equation}
\begin{equation}
	\begin{aligned}
	\ell_{fm}  = 1- \frac{(1+\beta^2) \cdot P (\mathbf{S}, \mathbf{G}) \cdot R((\mathbf{S}, \mathbf{G}))}
	{\beta^2 \cdot P(\mathbf{S}, \mathbf{G}) + R(\mathbf{S}, \mathbf{G})},
	\label{Fmloss}
	\end{aligned}
\end{equation}
where $\mathbf{G}(i)\in $\{0,1\} and $\mathbf{S}(i) \in$ [0, 1] are the ground truth label and predicted saliency score of the $i$-th pixel, respectively, 
$\beta^2$ is 0.3,
$P = \frac{TP} {TP+FP}$,
$R = \frac{TP} {TP+FN}$,
$TP(\mathbf{S}, \mathbf{G}) = \sum\nolimits_{i=1}^{W \cdot H} \mathbf{S}(i) \cdot \mathbf{G}(i)$,
$FP(\mathbf{S}, \mathbf{G}) = \sum\nolimits_{i=1}^{W \cdot H} \mathbf{S}(i) \cdot (1 - \mathbf{G}(i) )$,
and $FN(\mathbf{S}, \mathbf{G}) = \sum\nolimits_{i=1}^{W \cdot H} ( 1 - \mathbf{S}(i) ) \cdot \mathbf{G}(i)$.
Our comprehensive loss function helps our MCCNet better adapt to the special scenes of optical RSIs.

Besides, in the $t$-{th} MCCM, we generate the edge map $\bm{a}^{t}_\mathrm{e}$ by a model learned from minimizing with the edge loss $\mathbb{L}^{t}_{\mathrm{e}}$, which can be formulated as:
\begin{equation}
   \begin{aligned}
    \mathbb{L}^{t}_{\mathrm{e}}  =  \ell_{bce} (\mathrm{up}(\bm{a}^{t}_\mathrm{e}),\mathbf{G}_\mathrm{e})  ,
    \label{eq:EdgeLoss}
    \end{aligned}
\end{equation}
where $\mathbf{G}_\mathrm{e}$ is the ground truth of edge, generated in the same way as~\cite{2019EGNet}.
Therefore, the total loss $\mathbb{L}_{\mathrm{total}}$ of our MCCNet in the training phase can be expressed as:
\begin{equation}
   \begin{aligned}
    \mathbb{L}_{\mathrm{total}}  =  \sum\nolimits_{t=1}^5 (\mathbb{L}^{t}_{\mathrm{s}} + \mathbb{L}^{t}_{\mathrm{e}} ).
    \label{eq:TotalLoss}
    \end{aligned}
\end{equation}
\begin{table*}[t!]
  \centering
  \small
  \renewcommand{\arraystretch}{1.5}
  \renewcommand{\tabcolsep}{0.8mm}
  \caption{
    Quantitative results on two RSI-SOD datasets, including EORSSD and ORSSD.
    There are 23 state-of-the-art methods, including five traditional NSI-SOD methods, twelve CNN-based NSI-SOD methods, three traditional RSI-SOD methods, and three CNN-based RSI-SOD methods.
    $\uparrow$/$\downarrow$ means a larger/smaller score is better.
    The top three results are highlighted in \textcolor{red}{\textbf{red}}, \textcolor{blue}{\textbf{blue}} and \textcolor{green}{\textbf{green}}, respectively.
    }
\label{table:QuantitativeResults}
  
    \resizebox{1\textwidth}{!}{
\begin{tabular}{r|c|c|c|c|cccccccc|cccccccc}
\midrule[1pt]    
 \multirow{2}{*}{\normalsize{Methods}}
 & \multirow{2}{*}{\normalsize{Type}}
& Speed
& \#Param
& FLOPs
 & \multicolumn{8}{c|}{EORSSD~\cite{2021DAFNet}} 
 & \multicolumn{8}{c}{ORSSD~\cite{2019LVNet}}  \\
 
 \cline{6-13} \cline{14-21} 
         &  & (\emph{fps})$\uparrow$ & (M)$\downarrow$ & (G)$\downarrow$ & $S_{\alpha}\uparrow$ & $F_{\beta}^{\rm{max}}\uparrow$ & $F_{\beta}^{\rm{mean}}\uparrow$ & $F_{\beta}^{\rm{adp}}\uparrow$ & $E_{\xi}^{\rm{max}}\uparrow$ & $E_{\xi}^{\rm{mean}}\uparrow$ & $E_{\xi}^{\rm{adp}}\uparrow$ & $ \mathcal{M}\downarrow$
   	          & $S_{\alpha}\uparrow$ & $F_{\beta}^{\rm{max}}\uparrow$ & $F_{\beta}^{\rm{mean}}\uparrow$ & $F_{\beta}^{\rm{adp}}\uparrow$ & $E_{\xi}^{\rm{max}}\uparrow$ & $E_{\xi}^{\rm{mean}}\uparrow$ & $E_{\xi}^{\rm{adp}}\uparrow$ & $ \mathcal{M}\downarrow$\\
	     
\midrule[1pt]
RRWR$_{15}$~\cite{2015RRWR}  & T.N. & 0.3 & -  & - & .5992 & .3993 & .3686 & .3344 & .6894 & .5943 & .5639 & .1677
								         & .6835 & .5590 & .5125 & .4874 & .7649 & .7017 & .6949 & .1324  \\
HDCT$_{16}$~\cite{2016HDCT}    & T.N. & 7    &  -  & - & .5971 & .5407 & .4018 & .2658 & .7861 & .6376 & .5192 & .1088
									& .6197 & .5257 & .4235 & .3722 & .7719 & .6495 & .6291 & .1309  \\
DSG$_{17}$~\cite{2017DSG}        & T.N. & 0.6 & -  & - & .6420 & .5232 & .4597 & .4012 & .7260 & .6594 & .6188 & .1246
									& .7195 & .6238 & .5747 & .5657 & .7912 & .7337 & .7532 & .1041  \\
SMD$_{17}$~\cite{2017SMD}        & T.N. &  -  &  -  & - & .7101 & .5884 & .5473 & .4081 & .7697 & .7286 & .6416 & .0771
								        & .7640 & .6692 & .6214 & .5568 & .8230 & .7745 & .7682 & .0715  \\
RCRR$_{18}$~\cite{2018RCRR}   & T.N. & 0.3 & -  & - & .6007 & .3995 & .3685 & .3347 & .6882 & .5946 & .5636 & .1644
									& .6849 & .5591 & .5126 & .4876 & .7651 & .7021 & .6950 & .1277  \\
\hline
DSS$_{17}$~\cite{2017DSS}         	& C.N. & 8 & 62.23 & 114.6 & .7868 & .6849 & .5801 & .4597 & .9186 & .7631 & .6933 & .0186 
									 & .8262 & .7467 & .6962 & .6206 & .8860 & .8362 & .8085 & .0363 \\
RADF$_{18}$~\cite{2018RADF}    	& C.N. & 7 & 62.54 & 214.2 & .8179 & .7446 & .6582 & .4933 & .9140 & .8567 & .7162 & .0168 
									 & .8259 & .7619 & .6856 & .5730 & .9130 & .8298 & .7678 & .0382 \\
R3Net$_{18}$~\cite{2018R3Net}   	& C.N. & 2 & 56.16 &  47.5  & .8184 & .7498 & .6302 & .4165 & .9483 & .8294 & .6462 & .0171
									 & .8141 & .7456 & .7383 & .7379 & .8913 & .8681 & .8887 &  .0399\\
EGNet$_{19}$~\cite{2019EGNet}  	& C.N. & 9 & 108.07 & 291.9 & .8601 & .7880 & .6967 & .5379 & .9570 & .8775 & .7566 & .0110  
									 & .8721 & .8332 & .7500 & .6452 & .9731 & .9013 & .8226 & .0216 \\
PoolNet$_{19}$~\cite{2019PoolNet}  & C.N. & 25 & 53.63 & 123.4 & .8207 & .7545 & .6406 & .4611 & .9292 & .8193 & .6836 & .0210
									    & .8403 & .7706 & .6999 & .6166 & .9343 & .8650 & .8124 & .0358 \\
GCPA$_{20}$~\cite{2020GCPA}  	& C.N. & 23 & 67.06  & 54.3 & .8869 & .8347 & .7905 & .6723 & .9524 & .9167 & .8647 & .0102  
									 & .9026 & .8687 & .8433 & .7861 & .9509 & .9341 & .9205 & .0168 \\
ITSD$_{20}$~\cite{2020ITSD}  		& C.N. & 16 & 17.08 & 54.5 & .9050 & .8523 & .8221 & .7421 & .9556 & .9407 & .9103 & .0106
									   & .9050 & .8735 & .8502 & .8068 & .9601 & .9482 & .9335 & .0165 \\
MINet$_{20}$~\cite{2020MINet}  	& C.N. & 12 & 47.56 & 146.3 & .9040 & .8344 & .8174 & .7705 & .9442 & .9346 & .9243 & .0093
									   & .9040 & .8761 & .8574 & .8251 & .9545 & .9454 & .9423 & .0144 \\
GateNet$_{20}$~\cite{2020GateNet} & C.N. & 25 & 100.02 & 108.3 & .9114 & .8566 & .8228 & .7109 & .9610 & .9385 & .8909 & .0095
									    & .9186 & .8871 & .8679 & .8229 & .9664 & .9538 & .9428 & .0137 \\
U2Net$_{20}$~\cite{2020U2Net}  	& C.N. & 25  & 44.01 & 376.2 & \textcolor{green}{\textbf{.9199}} & \textcolor{blue}{\textbf{.8732}} & .8329 & .7221 & .9649 & .9373 & .8989 & \textcolor{green}{\textbf{.0076}}
								 	   & .9162 & .8738 & .8492 & .8038 & .9539 & .9387 & .9326 & .0166 \\
SUCA$_{21}$~\cite{2021SUCA}  	& C.N. & 24 & 117.71 & 56.4 & .8988 & .8229 & .7949 & .7260 & .9520 & .9277 & .9082 & .0097
									   & .8989 & .8484 & .8237 & .7748 & .9584 & .9400 & .9194 & .0145 \\
PA-KRN$_{21}$~\cite{2021PAKRN}  & C.N. & 16 & 141.06 & 617.7 & .9192 & .8639 & \textcolor{green}{\textbf{.8358}} & \textcolor{blue}{\textbf{.7993}} & .9616 & \textcolor{green}{\textbf{.9536}} &  \textcolor{green}{\textbf{.9416}} & .0104
									   & \textcolor{green}{\textbf{.9239}} & .8890 & \textcolor{green}{\textbf{.8727}} & \textcolor{green}{\textbf{.8548}} & .9680 & \textcolor{green}{\textbf{.9620}} & \textcolor{green}{\textbf{.9579}} & .0139 \\									   
\hline
VOS$_{18}$~\cite{2018VOS}  	  & T.R. & - & -   & - & .5082 & .2765 & .2107 & .1836 & .5982 & .4886 & .4767 & .2096
								  & .5366 & .3471 & .2717 & .2633 & .6514 & .5352 & .5826 & .2151 \\
CMC$_{19}$~\cite{2019CMC}  	  & T.R. & - & -   & - & .5798 & .3268 & .2692 & .2007 & .6803 & 5894 & .4890 & .1057
								 & .6033 & .3913 & .3454 & .3108 & .7064 & .6417 & .5996 & .1267 \\
SMFF$_{19}$~\cite{2019SMFF} & T.R. & - & -   & - & .5401 & .5176 & .2992 & .2083 & .7744 & .5197 & .5014 & .1434
								 & .5312 & .4417 & .2684 & .2496 & .7402 & .4920 & .5676 & .1854 \\
\hline
LVNet$_{19}$~\cite{2019LVNet}  	  & C.R. & 1.4 & -   & - & .8630 & .7794 & .7328 & .6284 & .9254 & .8801 & .8445 & .0146 
									      & .8815 & .8263 & .7995 & .7506 & .9456 & .9259 & .9195 & .0207\\
DAFNet$_{21}$~\cite{2021DAFNet}    & C.R. & 26 & 29.35 & 68.5 & .9166 & .8614 & .7845 & .6427 &  \textcolor{red}{\textbf{.9861}} & .9291 & .8446 & \textcolor{red}{\textbf{.0060}} 
									      & .9191 & \textcolor{green}{\textbf{.8928}} & .8511 & .7876 & \textcolor{blue}{\textbf{.9771}} & .9539 & .9360 & \textcolor{green}{\textbf{.0113}} \\ 
EMFINet$_{21}$~\cite{2021EMFINet} & C.R. & 25 & 107.26  & 480.9 & \textcolor{blue}{\textbf{.9290}} & \textcolor{green}{\textbf{.8720}} & \textcolor{blue}{\textbf{.8486}} & \textcolor{green}{\textbf{.7984}} & \textcolor{green}{\textbf{.9711}} & \textcolor{blue}{\textbf{.9604}} & \textcolor{blue}{\textbf{.9501}} & .0084
									     & \textcolor{blue}{\textbf{.9366}} & \textcolor{blue}{\textbf{.9002}} & \textcolor{blue}{\textbf{.8856}} & \textcolor{blue}{\textbf{.8617}} & \textcolor{green}{\textbf{.9737}} & \textcolor{blue}{\textbf{.9671}} & \textcolor{blue}{\textbf{.9663}} & \textcolor{blue}{\textbf{.0109}}  \\

\hline
\hline
\textbf{Ours} 					& C.R. & 95 & 67.65  & 112.8 & \textcolor{red}{\textbf{.9327}} & \textcolor{red}{\textbf{.8904}} & \textcolor{red}{\textbf{.8604}} & \textcolor{red}{\textbf{.8137}} & \textcolor{blue}{\textbf{.9755}} & \textcolor{red}{\textbf{.9685}} & \textcolor{red}{\textbf{.9538}} & \textcolor{blue}{\textbf{.0066}}
				       & \textcolor{red}{\textbf{.9437}} & \textcolor{red}{\textbf{.9155}} & \textcolor{red}{\textbf{.9054}} & \textcolor{red}{\textbf{.8957}} & \textcolor{red}{\textbf{.9800}} & \textcolor{red}{\textbf{.9758}} & \textcolor{red}{\textbf{.9735}} & \textcolor{red}{\textbf{.0087}}
\\
\toprule[1pt]
\multicolumn{19}{l}{\small{T.N.: Traditional NSI-SOD method; C.N.: CNN-based NSI-SOD method; T.R.: Traditional RSI-SOD method; C.R.: CNN-based RSI-SOD method. }} \\
\end{tabular}
}
\end{table*}

\begin{figure*}[t!]
\centering
\footnotesize
  \begin{overpic}[width=0.95\textwidth]{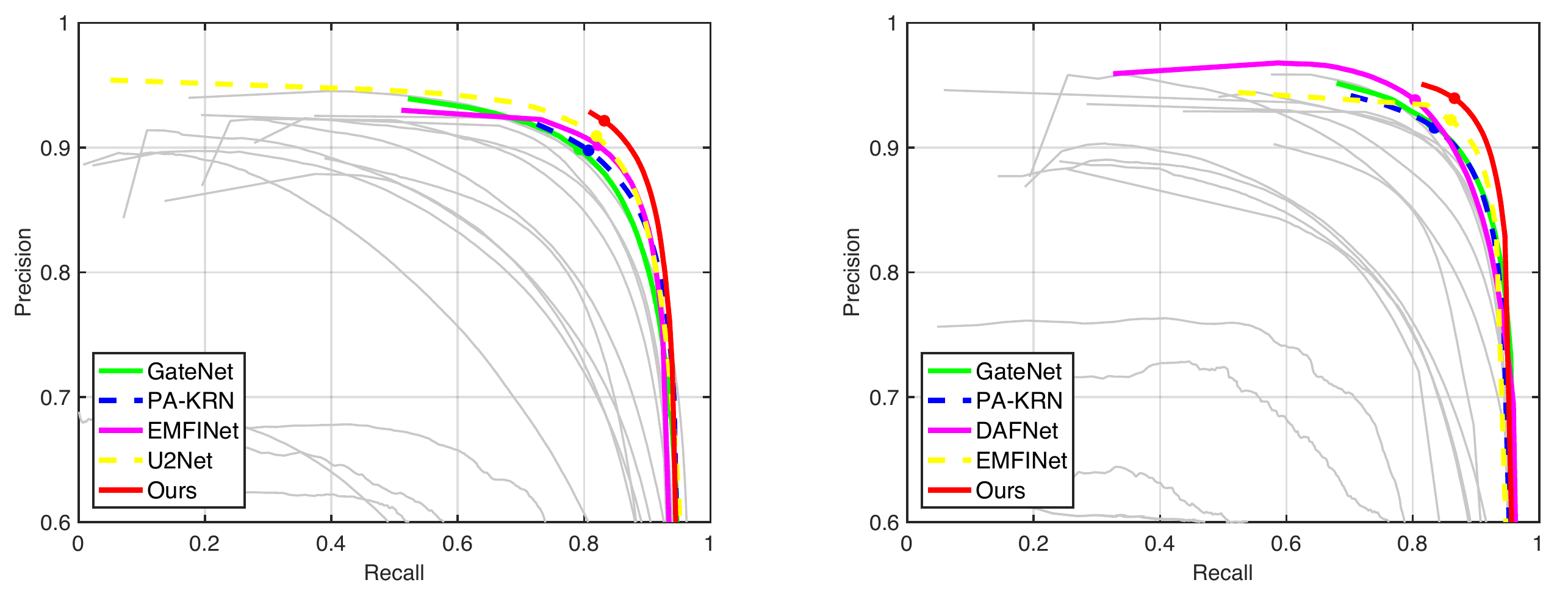}
    \put(19.18,-1.0){ (a) EORSSD~\cite{2021DAFNet} }
    \put(72.4,-1.0){ (b) ORSSD~\cite{2019LVNet} }   
  \end{overpic}
\caption{
Quantitative comparison in terms of PR curve on two datasets for RSI-SOD, \ie EORSSD and ORSSD.
We show the top five methods in color.
}
\label{PR_comparison}
\end{figure*}

\section{Experiments}
\label{sec:exp}

\subsection{Experimental Protocol}
\label{sec:ExpProtocol}
\textit{1) Datasets.}
We train and evaluate our method and the compared methods on two public RSI-SOD datasets.

\textbf{ORSSD}~\cite{2019LVNet} consists of 800 optical RSIs with corresponding pixel-wise ground truths,
including multiple scenes, such as ships, cars, airplanes, playgrounds, rivers, and islands.
We adopt 600 images with their ground truths for training and the remaining 200 images for testing.

\textbf{EORSSD}~\cite{2021DAFNet} expands the ORSSD dataset to include more complex and challenging scenes,
resulting in 2,000 optical RSIs with corresponding pixel-wise annotation,
which is the largest available RSI-SOD dataset.
We adopt 1,400 images with their ground truths for training and the remaining 600 images for testing.

\textit{2) Implementation Details.}
We conduct experiments of our proposed MCCNet on PyTorch~\cite{PyTorch} platform with an NVIDIA Titan X GPU (12GB memory).
During the network training, each optical RSI is resized to 256$\times$256 and augmented by flipping and rotation,
producing seven additional training samples.
We initialize the encoder network of our MCCNet and other newly added convolutional layers by the pre-trained VGG-16 model~\cite{2014VGG16ICLR}
and the normal distribution~\cite{InitialWei}, respectively.
We utilize the Adam optimizer~\cite{Adam} for network optimization with the batch size 8 and the initial learning rate $1e^{-4}$,
which will be divided by 10 after 30 epochs.
Following~\cite{2021DAFNet,2021EMFINet}, for the EORSSD dataset~\cite{2021DAFNet}, we train our MCCNet with 1,400 original optical RSI and GT pairs
and their 9,800 augmented samples for 39 epochs.
While for the ORSSD dataset~\cite{2019LVNet}, we train our MCCNet with 600 original optical RSI and GT pairs
and their 4,200 augmented samples for 34 epochs.

\textit{3) Evaluation Metrics.}
We use five evaluation metrics to evaluate our method and other compared methods:
$\textbf{S-measure}$ ($S_{\alpha}$, $\alpha$ = 0.5)~\cite{Fan2017Smeasure} is responsible for evaluating the structural similarity at object-aware and region-aware levels.
$\textbf{F-measure}$ ($F_{\beta}$)~\cite{Fmeasure} is the weighted harmonic average of precision and recall.
We set $\beta^2$ to 0.3 to emphasize the precision over recall, and adopt its maximum, mean and adaptive forms (\ie $F_{\beta}^{\rm{max}}$, $F_{\beta}^{\rm{mean}}$ and $F_{\beta}^{\rm{adp}}$) for comprehensive measure.
$\textbf{E-measure}$ ($E_{\xi}$)~\cite{Fan2018Emeasure} simultaneously captures the local match information at pixel-level and the global statistics at image-level.
We report its maximum, mean and adaptive values (\ie $E_{\xi}^{\rm{max}}$, $E_{\xi}^{\rm{mean}}$ and $E_{\xi}^{\rm{adp}}$).
$\textbf{Mean Absolute Error}$ (MAE, $\mathcal{M}$) evaluates the average pixel-level difference.
$\textbf{Precision-Recall curve}$ (PR) plots different combinations of precision and recall with the threshold ranging from 0 to 255.
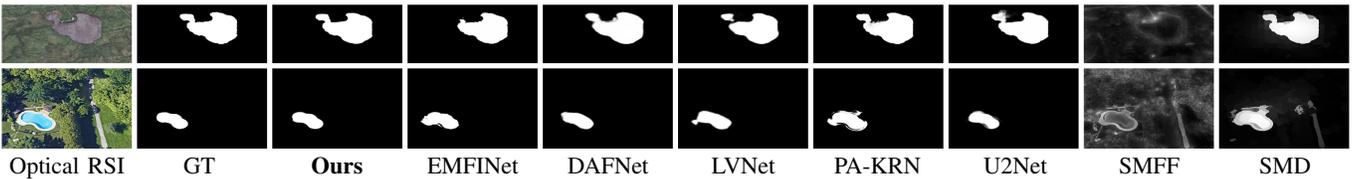
\begin{figure*}[t!]
    \centering
    \small
	\begin{overpic}[width=1\textwidth]{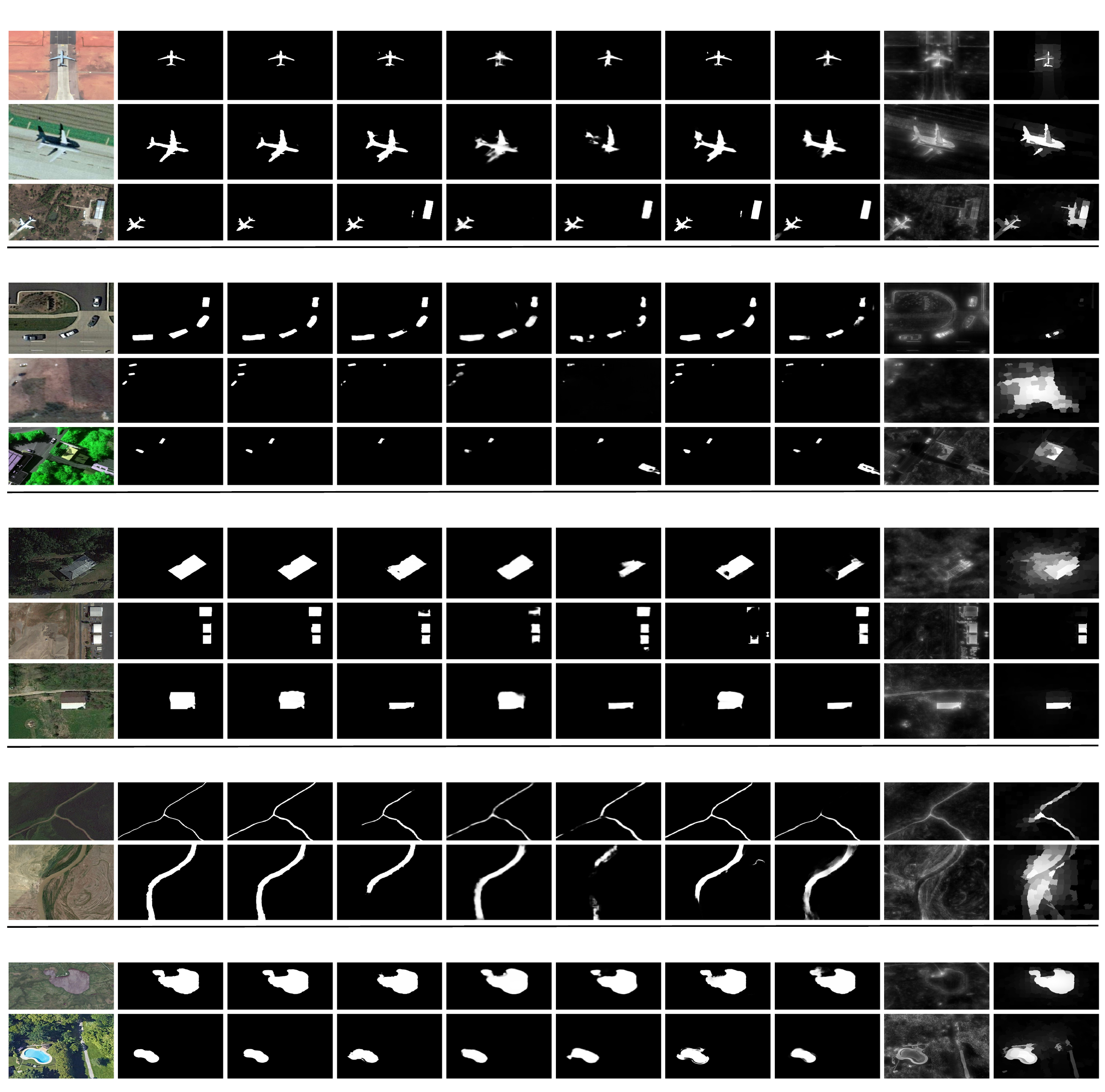}

    \put(0.45,96.9){ Airplane: General airplane / Airplane with shadows / Airplane with interferences}
    \put(0.45,74.1){ Car: Multiple cars / Multiple tiny cars / Cars with complex background}
    \put(0.45,51.9){ Building: General building / Multiple buildings / Building with inconsistent colors}
    \put(0.45,28.9){ River: River with irregular topology / River with low contrast}
    \put(0.45,12.6){ Pool: Pool with complex geometry / Pool with interferences}

    \put(0.67,-0.65){ Optical RSI}
    \put(13.45,-0.65){ GT}
    \put(22.8,-0.65){ \textbf{Ours}}
    \put(31.3,-0.65){ EMFINet }
    \put(41.6,-0.65){ DAFNet }
    \put(52.15,-0.65){ LVNet }
    \put(61.15,-0.65){ PA-KRN }
    \put(72.1,-0.65){ U2Net }
    \put(82,-0.65){ SMFF }
    \put(92.3,-0.65){ SMD } 
    
    \end{overpic}
	\caption{Visual comparisons with seven representative state-of-the-art methods,
	including three CNN-based RSI-SOD methods (EMFINet~\cite{2021EMFINet}, DAFNet~\cite{2021DAFNet} and LVNet~\cite{2019LVNet}),
	two CNN-based NSI-SOD methods (PA-KRN~\cite{2021PAKRN} and U2Net~\cite{2020U2Net}),
	one traditional RSI-SOD method (SMFF~\cite{2019SMFF}),
	and one traditional NSI-SOD method (SMD~\cite{2017SMD}), on various scenes.
	Please zoom-in for the best view.
    }
    \label{fig:VisualExample}
\end{figure*}

\subsection{Comparison with State-of-the-art Methods}
We compare our proposed method with 23 state-of-the-art SOD methods, which can be divided into four categories.
The first one contains the traditional NSI-SOD methods, including RRWR~\cite{2015RRWR}, HDCT~\cite{2016HDCT}, DSG~\cite{2017DSG}, SMD~\cite{2017SMD}, and RCRR~\cite{2018RCRR}.
The second one includes the CNN-based NSI-SOD methods, including DSS~\cite{2017DSS}, RADF~\cite{2018RADF}, R3Net~\cite{2018R3Net}, EGNet~\cite{2019EGNet}, PoolNet~\cite{2019PoolNet}, GCPA~\cite{2020GCPA}, ITSD~\cite{2020ITSD}, MINet~\cite{2020MINet}, GateNet~\cite{2020GateNet}, U2Net~\cite{2020U2Net}, SUCA~\cite{2021SUCA}, and PA-KRN~\cite{2021PAKRN}.
The third one includes the traditional RSI-SOD methods, including VOS~\cite{2018VOS}, CMC~\cite{2019CMC}, and SMFF~\cite{2019SMFF}.
The last one contains the CNN-based RSI-SOD methods, including LVNet~\cite{2019LVNet}, DAFNet~\cite{2021DAFNet}, and EMFINet~\cite{2021EMFINet}.
For a fair comparison, we use the saliency maps provided by the available public RSI-SOD benchmarks~\cite{2019LVNet,2021DAFNet} and/or by the authors.
Specifically, we retrain seven recent CNN-based NSI-SOD methods, \ie GCPA, ITSD, MINet, GateNet, U2Net, SUCA and PA-KRN, on EORSSD and ORSSD datasets with their default settings using the same training data as our method.
\textit{1) Quantitative Comparison.}
In Tab.~\ref{table:QuantitativeResults}, we report the quantitative comparison results of our method and all compared methods on two RSI-SOD datasets in terms of $S_{\alpha}$, $F_{\beta}^{\rm{max}}$, $F_{\beta}^{\rm{mean}}$, $F_{\beta}^{\rm{adp}}$, $E_{\xi}^{\rm{max}}$, $E_{\xi}^{\rm{mean}}$, $E_{\xi}^{\rm{adp}}$, and $\mathcal{M}$, among which the higher the first seven evaluation metrics, the better, while the last one is opposite to them.

Overall, our method shows excellent performance as compared with all four categories of methods on EORSSD and ORSSD.
Specifically, on the EORSSD dataset, our method is weaker than DAFNet in terms of $F_{\beta}^{\rm{max}}$ and $\mathcal{M}$, but is significantly better than DAFNet on the other six metrics, \eg $F_{\beta}^{\rm{adp}}$: 0.8137 (Ours) \emph{v.s.} 0.6427 (DAFNet), and $E_{\xi}^{\rm{adp}}$: 0.9538 (Ours) \emph{v.s.} 0.8446 (DAFNet).
While on the ORSSD dataset, our method fully surpasses DAFNet in all eight metrics, \eg $S_{\alpha}$: 0.9437 (Ours) \emph{v.s.} 0.9191 (DAFNet), and $\mathcal{M}$: 0.0087 (Ours) \emph{v.s.} 0.0113 (DAFNet).
Compared with the latest CNN-based RSI-SOD method EMFINet, our method consistently outperforms it on both datasets, \eg $F_{\beta}^{\rm{max}}$: 2.11\% and 1.70\% better than it and $\mathcal{M}$: 21.43\% and 20.18\% lower than it on EORSSD and ORSSD, respectively.
In comparison to the eight traditional methods, including NSI- and RSI-SOD, our method is a lot ahead of them.
Even though the CNN-based NSI-SOD methods are retrained on optical RSI data, their performance is generally lower than that of specialized RSI-SOD methods, which illustrates the urgency and necessity of proposing specialized solutions.
In addition, we present PR curves in Fig.~\ref{PR_comparison}.
We observe that the specialized CNN-based methods show great strength, and the PR curve of our method is closer to the upper right corner than all compared methods, which is consistent with the remarkable quantitative results of our method on two datasets in Tab.~\ref{table:QuantitativeResults}.
%

\begin{figure*}[t!]
\centering
\footnotesize
  \begin{overpic}[width=1\textwidth]{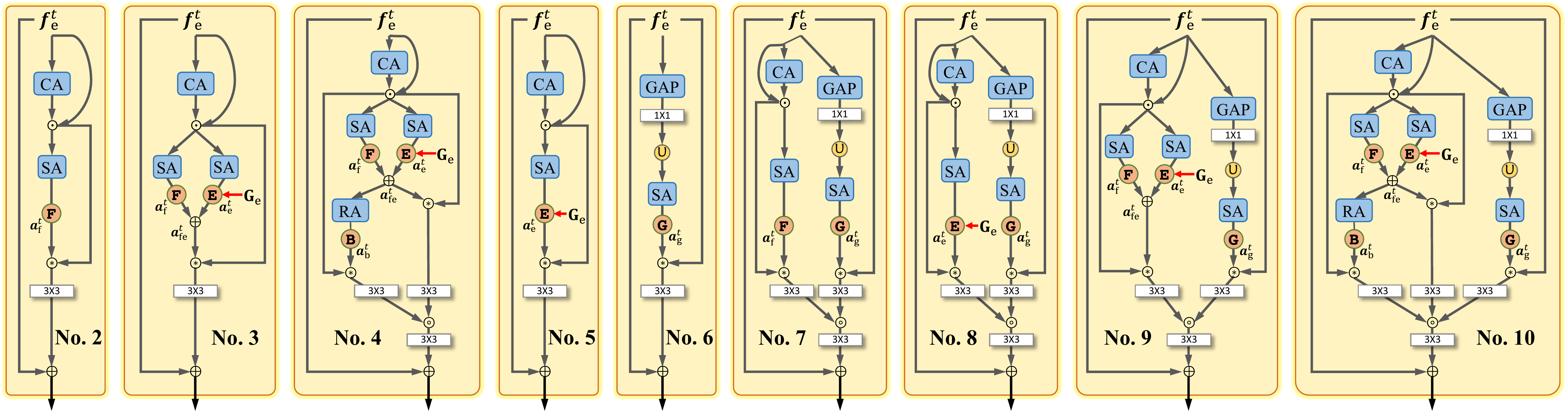}
  \end{overpic}
\caption{
Structures of eight MCCM variants (\ie No. 2 $\sim$ No. 9) and the full MCCM (\ie No. 10).
Please zoom-in for viewing details.
}
\label{MCCM_Ablation_study}
\end{figure*}

\textit{2) Computational Complexity Comparison.}
We measure the computational complexity from three aspects, including inference speed (without I/O time), network parameters and FLOPs, which are captured from the available public RSI-SOD benchmarks~\cite{2019LVNet,2021DAFNet} and our retraining, and report them in Tab.~\ref{table:QuantitativeResults}.
Overall, we can find that most CNN-based methods run in real-time (23$\sim$25 \emph{fps}), especially our method reaches an astonishing inference speed of 95 \emph{fps}, which is friendly to practical applications.
The network parameters and FLOPs of our method are at the midstream level.
However, compared with the second best method EMFINet, our network parameters and FLOPs are much smaller, \eg \#Param: 67.65M (Ours) \emph{v.s.} 107.26M (EMFINet), and FLOPs: 112.8M (Ours) \emph{v.s.} 480.9M (EMFINet).
From the above quantitative comparison and computational complexity comparison, we can conclude that our method is effective and efficient.

\textit{3) Visual Comparison.}
We show some representative visual examples of optical RSIs in Fig.~\ref{fig:VisualExample}, including airplane, car, building, river and pool.
We summarize the five specific scenes as follows:
1) three cases of airplane: general airplane, airplane with shadows, and airplane with interferences;
2) three cases of car: multiple cars, multiple tiny cars, and cars with complex background;
3) three cases of building: general building, multiple buildings, and building with inconsistent colors;
4) two cases of river: river with irregular topology and river with low contrast, and 
5) two cases of pool: pool with complex geometry and pool with interferences.
The above cases include challenging scenes of optical RSIs, such as multiple objects, tiny objects, object with interferences, object with shadows, complex background, low contrast and complex topology.
We present the saliency maps of the representative methods in four categories, including EMFINet, DAFNet, LVNet, PA-KRN, U2Net, SMFF, and SMD.

Obviously, the two traditional methods SMFF and SMD are often confused by the unique scenes of optical RSIs.
The two recent CNN-based NSI-SOD methods PA-KRN and U2Net often show weak inadaptability to optical RSIs data.
The three CNN-based RSI-SOD methods can overcome difficult scenes to a certain extent, but generate saliency maps with flaws.
Using the complementarity of four kinds of content, \ie foreground, edge, background, and global image-level content, our method can accurately locate salient objects and outline fine details, which shows strong adaptability and robustness in these scenes.

\begin{table}[!t]
\centering
\caption{Ablation study on evaluating the individual contribution of each content in MCCM.
  Baseline is the encoder-decoder network with skip connections.
  FG, EG, BG and GIC mean foreground, edge, background and global image-level content, respectively.
  The best result in each column is \textbf{bold}.
  }
\label{table:MCCM_Ablation_study}
\renewcommand{\arraystretch}{1.5}
\renewcommand{\tabcolsep}{0.8mm}
\begin{tabular}{c|ccccc||cc|cc}
\toprule

 \multirow{2}{*}{No.} & \multirow{2}{*}{Baseline} & \multirow{2}{*}{FG} & \multirow{2}{*}{EG}  & \multirow{2}{*}{BG} & \multirow{2}{*}{GIC}  
 & \multicolumn{2}{c|}{EORSSD~\cite{2021DAFNet}} 
 & \multicolumn{2}{c}{ORSSD~\cite{2019LVNet}} \\
 
 \cline{7-10}
    & & & & & 
    & $F_{\beta}^{\rm{max}}\uparrow$ & $E_{\xi}^{\rm{max}}\uparrow$
    & $F_{\beta}^{\rm{max}}\uparrow$ & $E_{\xi}^{\rm{max}}\uparrow$ \\
\midrule
1 & \Checkmark   &                     &                    &                     &                     &   .8687 & .9585 & .8878 & .9608  \\ 
2 &  \Checkmark & \Checkmark &                     &                     &                     &   .8829 & .9703 & .9043  & .9711  \\
3 &  \Checkmark & \Checkmark & \Checkmark &                     &                     &   .8856 & .9727 & .9066  & .9742   \\
4 &  \Checkmark & \Checkmark & \Checkmark & \Checkmark &                     &   .8870 & .9747 & .9096  & .9774   \\
\hline
5 &  \Checkmark &                & \Checkmark &                      &                     &   .8842 & .9713 & .9047  & .9718   \\
6 &  \Checkmark &                &                     &                      & \Checkmark &   .8740 & .9623 & .9014  & .9689   \\
7 &  \Checkmark & \Checkmark &                 &                     & \Checkmark &   .8855 & .9728 & .9116  & .9755   \\
8 &  \Checkmark &                 & \Checkmark &                     & \Checkmark &   .8864 & .9736 & .9114  & .9746   \\
9 &  \Checkmark & \Checkmark & \Checkmark &                 & \Checkmark &   .8896 & .9747 & .9127  & .9776   \\
\hline
10 &  \Checkmark & \Checkmark & \Checkmark & \Checkmark & \Checkmark & \bf{.8904} & \bf{.9755} & \bf{.9155} & \bf{.9800} \\
\toprule
\end{tabular}
\end{table}

\subsection{Ablation Studies}
\label{Ablation Studies}
Here, we conduct comprehensive experiments to evaluate the effectiveness of important components of our MCCNet on EORSSD and ORSSD datasets.
In particular, we investigate
1) the individual contribution of each content in MCCM,
2) the necessity of merging the original content in MCCM, and
3) the effectiveness of our comprehensive loss function. 

For each variant experiment, we rigorously retrain it with the same parameter settings and datasets as in Sec.~\ref{sec:ExpProtocol}.

\textit{1) The individual contribution of each content in MCCM}.
To evaluate the individual contribution of each content, \ie foreground (FG), edge (EG), background (BG) and global image-level content (GIC), in MCCM, we first provide four progressive variants of MCCM on the upper part of Tab.~\ref{table:MCCM_Ablation_study}:
1) Baseline, which is the encoder-decoder network with skip connections (\ie replacing MCCM with skip connection in MCCNet),
2) Baseline+FG,
3) Baseline+FG+EG, and
4) Baseline+FG+EG+BG.
%
For an intuitive understanding of variants, we illustrate three variants (No. 2 $\sim$ No. 4) in Fig.~\ref{MCCM_Ablation_study}.

Based on the quantitative results in Tab.~\ref{table:MCCM_Ablation_study}, we observe consistently the increasing trends of performance on both datasets.
FG greatly activates the potential of the network, and based on it, by additionally exploring the complementation of EG and BG, the performance takes off further. 
Notably, although GIC only provides the specific basic tone of source features, as expressed in Sec.~\ref{sec:MCCM} and Fig.~\ref{feature_visual}, it increases $F^{\rm{max}}_{\beta}$ from 0.8879 to 0.8904 on the EORSSD dataset and from 0.9096 to 0.9155 on the ORSSD dataset.
In summary, our full MCCM improves ``Baseline" by 2.17\% and 1.70\% on $F^{\rm{max}}_{\beta}$ and $E^{\rm{max}}_{\xi}$, respectively, on the EORSSD dataset.
The performance improvement is more significant on the ORSSD dataset, that is, our full MCCM improves ``Baseline" by 2.77\% and 1.92\% on $F^{\rm{max}}_{\beta}$ and $E^{\rm{max}}_{\xi}$, respectively.
%

Moreover, to comprehensively analyze the different combinations of four kinds of content, we provide another five variants of MCCM on the middle part of Tab.~\ref{table:MCCM_Ablation_study}:
5) Baseline+EG,
6) Baseline+GIC,
7) Baseline+FG+GIC,
8) Baseline+EG+GIC, and
9) Baseline+FG+EG+GIC.
We also illustrate these five variants (No. 5 $\sim$ No. 9) in Fig.~\ref{MCCM_Ablation_study}.
Notably, since BG is jointly defined by FG and EG according to Eq.~\ref{eq:5}, BG cannot be obtained by the single FG or the single EG.

By comparing the fifth and sixth variants with the first one respectively, we can clearly observe the contributions of EG and GIC.
By comparing the seventh and eighth variants with the sixth one respectively, we can find that based on GIC, FE or EG can further promote the performance.
When we abandon BG in MCCM, the performance of ``Baseline+FG+EG+GIC" is adversely affected.
Through the comprehensive comparison of all nine variants and our complete MCCM, we can conclude that each content in MCCM contributes to the final excellent performance and the proposed MCCM is effective.

\begin{table}[!t]
\centering
\caption{
Ablation study on proving the necessity of merging the original content in MCCM.
  The best result in each column is \textbf{bold}.
  }
\label{AblationOriginalContent}
\renewcommand{\arraystretch}{1.5}
\renewcommand{\tabcolsep}{1.2mm}
\begin{tabular}{c||cc|cc}
\toprule

\multirow{2}{*}{Models}
 & \multicolumn{2}{c|}{EORSSD~\cite{2021DAFNet}}
 & \multicolumn{2}{c}{ORSSD~\cite{2019LVNet}} \\
 
    & $F_{\beta}^{\rm{max}}\uparrow$ & $E_{\xi}^{\rm{max}}\uparrow$
    & $F_{\beta}^{\rm{max}}\uparrow$ & $E_{\xi}^{\rm{max}}\uparrow$ \\
\midrule
\textit{w/o original content} &   .8859 & .9729 & .9120 & .9763   \\ 

\hline
\textit{w/ original content} (\textbf{Ours}) &  \textbf{.8904} & \textbf{.9755} & \textbf{.9155} & \textbf{.9800} \\
\toprule
\end{tabular}
\end{table}

\textit{2) The necessity of merging the original content in MCCM.}
To prove the necessity of merging the original content in MCCM, we provide a variant that removes the original content in Eq.~\ref{eq:9}, \ie \textit{w/o original content}.
As shown in Tab.~\ref{AblationOriginalContent}, we can observe that the performance degradation occurs in \textit{w/o original content}, \eg $F^{\rm{max}}_{\beta}$ and $E^{\rm{max}}_{\xi}$ are reduced by 0.0045 and 0.0026 on the EORSSD dataset, and 0.0035 and 0.0037 on the ORSSD dataset.
This demonstrates that the basic information of salient regions provided by the original content is necessary for a better performance.

\textit{3) The effectiveness of our comprehensive loss function.}
To validate the effectiveness of our comprehensive loss function, we provide three loss variants for network training:
1) the single BCE loss,
2) BCE loss with IoU loss, and
3) BCE loss with F-m loss.
%
We report the quantitative results in Tab.~\ref{AblationStudyLoss}.

Our MCCNet trained with the single BCE loss achieves acceptable performance, \eg $F^{\rm{max}}_{\beta}$: 0.8747 and $E^{\rm{max}}_{\xi}$: 0.9705 on the EORSSD dataset, and $F^{\rm{max}}_{\beta}$: 0.9094 and $E^{\rm{max}}_{\xi}$: 0.9728 on the ORSSD dataset.
And with the assistance of IoU loss or F-m loss, the performance is further improved by about 0.0018$\sim$0.0127 in $F^{\rm{max}}_{\beta}$ and 0.0018$\sim$0.0038 in $E^{\rm{max}}_{\xi}$.
Integrating the three losses together to train our MCCNet, our MCCNet achieves the best performance, which increases the simplest variant by 0.0158 in $F^{\rm{max}}_{\beta}$ on the EORSSD dataset and 0.0072 in $E^{\rm{max}}_{\xi}$ on the ORSSD dataset.
In general, the above analysis clearly verifies the effectiveness of our comprehensive loss function, and training network with suitable losses is efficient to boost our method without additional parameters.

\begin{table}[!t]
\centering
\caption{Ablation study on evaluating the effectiveness of our comprehensive loss function.
  BCE, IoU and F-m represent BCE loss, IoU loss and F-measure loss, respectively.
  The best result in each column is \textbf{bold}.
  }
\label{AblationStudyLoss}
\renewcommand{\arraystretch}{1.5}
\renewcommand{\tabcolsep}{1.2mm}
\begin{tabular}{c|ccc||cc|cc}
\toprule

 \multirow{2}{*}{No.} & \multirow{2}{*}{BCE}  & \multirow{2}{*}{IoU}  & \multirow{2}{*}{F-m}  
 & \multicolumn{2}{c|}{EORSSD~\cite{2021DAFNet}} 
 & \multicolumn{2}{c}{ORSSD~\cite{2019LVNet}} \\
 
 \cline{5-8}
    & & &
    & $F_{\beta}^{\rm{max}}\uparrow$ & $E_{\xi}^{\rm{max}}\uparrow$
    & $F_{\beta}^{\rm{max}}\uparrow$ & $E_{\xi}^{\rm{max}}\uparrow$ \\
\midrule
1 & \Checkmark   &                     &                      &   .8746 & .9705 & .9094 & .9728   \\ 
2 &  \Checkmark  & \Checkmark &                      &   .8873 & .9726 & .9136  & .9766  \\
3 &  \Checkmark  & 			  & \Checkmark &   .8857 & .9723 & .9112  & .9760  \\

\hline
4 &  \Checkmark & \Checkmark & \Checkmark  & \bf{.8904} & \bf{.9755} & \bf{.9155} & \bf{.9800} \\
\toprule
\end{tabular}
\end{table}

\section{Conclusion}
\label{sec:con}
In this paper, we propose an effective Multi-Content Complementation Module to model the complementarity of multiple content, including foreground, edge, background, and global image-level content, for optical RSI data.
In MCCM, the foreground map and edge map are directly integrated to complete salient regions, and then each content complements others in an adaptive manner.
Moreover, we equip MCCM to the encoder-decoder network, and propose a full solution, namely Multi-Content Complementation Network, for RSI-SOD.
MCCMs on five feature sizes can highlight salient regions well, and successfully address the variable object scales/types/quantities of optical RSI data.
Finally, a comprehensive loss function is employed in the training phase to boost performance.
We conduct extensive experiments on two public RSI-SOD datasets.
The experimental results demonstrate the superiority of the proposed MCCNet as well as the effectiveness of the proposed MCCM.
Moreover, the fast inference speed of 95 \emph{fps} is extremely conducive to applying our MCCNet in practical applications.



\ifCLASSOPTIONcaptionsoff
  \newpage
\fi

\bibliographystyle{IEEEtran}
\bibliography{ORSIref}

%



%

\end{document}